\documentclass[a4paper,twoside]{article}

\usepackage{epsfig}
\usepackage{calc}
\usepackage{color}
\usepackage{amssymb}
\usepackage{amstext}
\usepackage{amsmath}
\usepackage{amsthm}
\usepackage{multicol}
\usepackage{pslatex}
\usepackage{apalike}
\usepackage{SCITEPRESS}
\usepackage[table,xcdraw]{xcolor}
\usepackage{textcomp}
\usepackage{subfig}

\begin{document}

\title{Convolutional Neural Networks: A Binocular Vision Perspective}

\author{\authorname{Yigit Oktar\sup{1}, Diclehan Karakaya\sup{2}, Oguzhan Ulucan\sup{2}, Mehmet Turkan\sup{2}}
\affiliation{\sup{1}Department of Computer Engineering, Izmir University of Economics, Izmir, Turkey}
\affiliation{\sup{2}Department of Electrical and Electronics Engineering, Izmir University of Economics, Izmir, Turkey}
}



\abstract{It is arguable that whether the single camera captured (monocular) image datasets are sufficient enough to train and test convolutional neural networks (CNNs) for imitating the biological neural network structures of the human brain. As human visual system works in binocular, the collaboration of the eyes with the two brain lobes needs more investigation for improvements in such CNN-based visual imagery analysis applications. It is indeed questionable that if respective visual fields of each eye and the associated brain lobes are responsible for different learning abilities of the same scene. There are such open questions in this field of research which need rigorous investigation in order to further understand the nature of the human visual system, hence improve the currently available deep learning applications. This position paper analyses a binocular CNNs architecture that is more analogous to the biological structure of the human visual system than the conventional deep learning techniques. While taking a structure called \textit{optic chiasma} into account, this architecture consists of basically two parallel CNN structures associated with each visual field and the brain lobe, fully connected later possibly as in the primary visual cortex (V1). Experimental results demonstrate that binocular learning of two different visual fields leads to better classification rates on average, when compared to classical CNN architectures.}

\onecolumn \maketitle \normalsize \vfill

\section{\uppercase{INTRODUCTION}}
\label{sec:introduction}

\noindent In the last decade, deep neural networks (deep learning) as improved versions of artificial neural networks (ANNs) have become very popular in academic and industrial applications of machine learning for object classification and recognition~\cite{agrawal2014analyzing}, (biomedical) image and video analysis~\cite{zhou2017fine}, speech and natural language processing, and face detection and recognition~\cite{sharifara2014general}. A specific and indeed very successful application of deep neural networks, namely convolutional neural networks (CNNs) are widely used and mostly applied to analyze visual contents~\cite{krizhevsky2012imagenet}. CNNs automatically extract hierarchical attributes (features) from visual datasets in large volumes with parallel calculation using GPUs, and there is huge amount of research going on for possible practical applications of CNNs together with serious industry investments, e.g., Facebook\textcopyright , Google\textcopyright, Amazon\textcopyright, Instagram\textcopyright.

CNNs and similar other deep learning architectures are at best analogous to the neural networks in the human brain, specifically to the visual system. In reality, it is not completely known how the human brain works; hence it is not possible to discuss that CNNs work exactly the same way as the neural networks in the human brain. However, there are certain parallels drawn in-between these two, and within particular constrained contexts analogy-based comparisons can be performed. Nevertheless, inspired by certain biological structures of the human brain, CNNs are subject to an optimization problem similar to most of the machine learning algorithms, and are particularly based on multi-layer perceptrons which require relatively less preprocessing when compared to other traditional methods~\cite{krizhevsky2012imagenet}.

A crucial point that needs a well-investigation is the interaction between visual fields of each eye and the brain. The information captured by the two eyes are transferred as electrical signals to the primary visual cortex through the primary visual pathway. Undoubtedly, the human visual system is binocular which helps reconstruct three-dimensional images of the sensed real-world scenes. In this way, life gets easy by means of neural learning in the brain, especially for estimating near-future possibilities in some time instants, and also classifying and recognizing partially or even completely occluded objects~\cite{humansee}. 

This position paper argues that whether the currently available single camera captured images are sufficient enough for training CNNs for imitating the neural network structures of the human brain, or are there some other important improvements still available but currently missing for a complete analysis? Are respective visual fields of each eye and the associated brain lobe responsible for different learning abilities of the same scene? What happens if there were two different but parallel CNN structures associated with each visual field and the brain lobe, fully-connected or concatenated as in the primary visual cortex? Is there a voting or averaging or some other (non)linear mechanism between the results obtained from these two different visual fields and brain lobes in order to produce a final outcome? These are open questions in this field of research which needs rigorous investigation in order to further understand the nature of the human visual system. The remaining parts of this paper discuss some of these questions inspired by biological evidences and provide possible future research directions of this domain of research.

\section{\uppercase{Human Visual System vs. Convolutional Neural Networks}}

\noindent Although the vision process is initialized in the eyes, the actual interpretation of what has been seen results in the primary visual cortex of the brain. The whole investigation starts with the research of Hubel and Wiesel in 1950s~\cite{wurtz2009recounting}. The visual cortexes of cats and monkeys are analyzed, and there are basically two differently activated visual neural cells named as simple (S) cells and complex (C) cells those respond to the visual environment. The S cells are activated to recognize basic shapes such as straight lines with a specific angle in a fixed region. The C cells on the other hand have larger \textit{receptive fields} and these cells are not sensitive to the specific position in the visual region. A receptive field is defined for a single visual neuron which activates (fires) that visual neuron. Every visual neuron cell have such typical receptive fields which partially overlap covering the entire visual field. In CNNs analogously, the convolution operation simulates the response of an individual visual neuron to its visual environment.

The deep and hierarchical structure of neural networks in the brain has an important function for classifying and recognizing different objects. CNNs are indeed simulating this deep and hierarchical structure such that the deep connections between neurons and hierarchical organizations are similar to those of the visual pathway. A fact that CNNs are somewhat simulating the visual pathway is that Gabor-like filters usually appear within~\cite{luan2018gabor}. It was claimed that Gabor filters might be helpful to model S cells in visual cortex of mammalian brains~\cite{daugman1985uncertainty} corresponding to convolutional layers in CNNs.

In addition, both linear and nonlinear (such as pooling) operations are carried out in the primary visual cortex~\cite{laskar2018correspondence}. In CNNs, subsampling or pooling layers simulate the function of C cells, providing a mechanism of translation and rotation invariance to a certain extent~\cite{maida2016cognitive}.

Biologically, the captured electrical signals are carried through axons, and the strength of connections are adjusted at synapses before the signals are received by the dendrites of the next neuron~\cite{eluyode2013comparative}. Then, an action potential is created by the neuron when its input signal strength exceeds a certain threshold. In order to simulate the nonlinear characteristics of this biological process, there are various types of activation functions used in CNNs. Rectified Linear Unit (ReLU) is commonly employed as an activation function in recent studies, because of its pace, good performance, effectiveness with multiple layer systems, usability with the back-propagation process. ReLU simply maps negative input values to zero. Some other variants of ReLU, such as Leaky ReLU and Exponential Linear Unit (ELU), avoid the vanishing gradient problem of RELU for negative inputs~\cite{clevert2015fast}.

In short, human visual system has been the core role model for current CNN architectures, and many other parallels can be drawn when both systems are investigated thoroughly. However, CNNs are currently not fully capable of simulating the human visual system and have certain limitations as mentioned in the remaining part of this section.

\subsection{Limitations of Convolutional Neural Networks}

\noindent A recently popularized weakness of CNNs is their vulnerability to adversarial examples, inputs that have been slightly altered to trick the system, drastically decreasing the accuracy in classifications tasks. As reported, the primary cause of this vulnerability is related to the linearity of neural networks~\cite{goodfellow2014explaining}. On the other hand, pooling (usually max-pooling is used) gives some sort of limited rotation and translation invariance to CNNs. However, standard CNNs do not possess rotation or more generally total affine transformation invariance. To handle this problem, CNNs either need exponentially large number of training samples covering all cases, or replication of feature detectors on a grid, which indeed grows exponentially with the number of dimensions~\cite{sabour2017dynamic}.
\begin{figure}[!t]
	\centering
	{\epsfig{file = 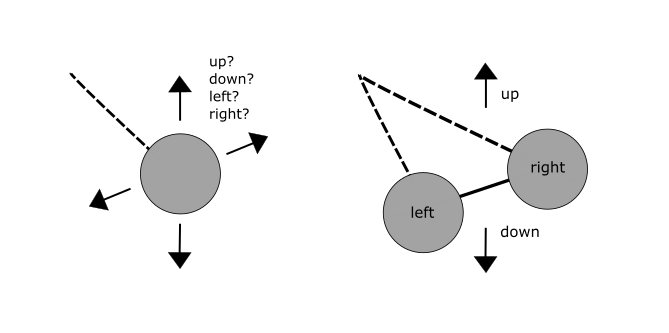, width = 8cm}}
	\caption{Considering a binocular CNNs structure may provide handedness and a sense of direction, in which each eye behaves as a reference for the other; thus fixing the sense of up, down, left, and right directions.}
	\label{fig:sag_sol}
\end{figure}

It is hypothesized that through a binocular architecture, some of the current drawbacks of CNNs can be eliminated. As reported, conventional monocular CNNs do not possess handedness. As depicted in Fig.~\ref{fig:sag_sol}, having a binocular structure may provide a sense of handedness in which each eye provides a reference point for the other; thus fixing the sense of up, down, left, and right directions. On the other side, monocular structures clearly will have difficulty of registering a default pose without a reference point. A monocular structure might also be responsible for inability of perceiving rotation and certain other affine transformations such as reflection. It can also be argued that monocular structures could be the cause of the susceptibility to certain adversarial attacks. Therefore, a binocular vision perspective of CNNs is to be discussed in the following section.

\section{\uppercase{A BINOCULAR VISION PERSPECTIVE}}

\begin{figure}[!t]
	\centering
	{\epsfig{file = 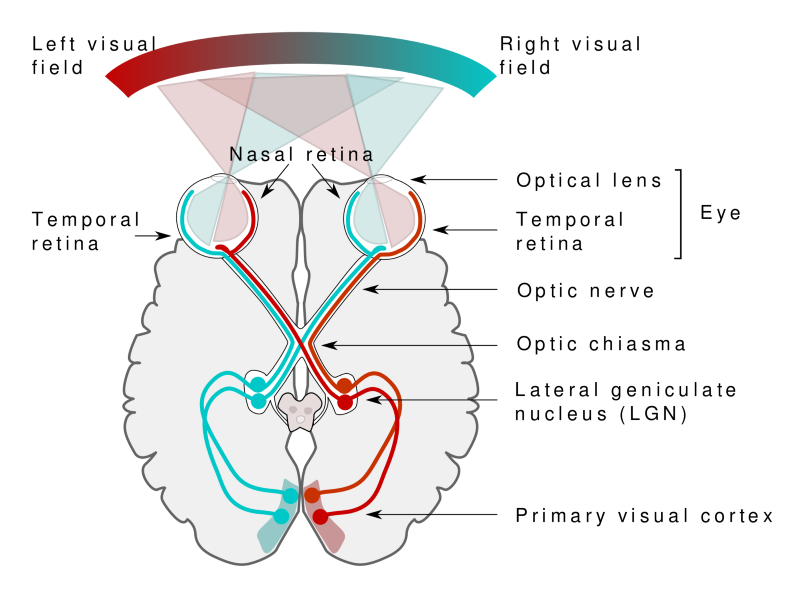, width = 7.5cm}}
	\caption{A simple illustration of the human visual pathway. Source: Miquel Perello Nieto (Creative Commons License)~\cite{hvp}}.
	\label{fig:example1}
\end{figure}

\noindent Undoubtedly, the human visual system is binocular. A simple schema of the human visual pathway is illustrated in Fig.~\ref{fig:example1}. Light signals received on the retina of each eye are transferred to lateral geniculate nucleus (LGN) through the optic chiasma via optic nerves. Then, the final processing occurs in the primary visual cortex.

This position paper mainly focuses on the function of optic chiasma, which is responsible for transmitting right visual fields to the left brain lobe and left visual fields to the right brain lobe. The advantages of such a structure and its absence will further be discussed in the following subsections.

\subsection{Proposed Idea vs. Related Work}

\noindent Binocular, or in general terms, double input consideration of neural networks is not a completely new concept, and goes back to $1994$ under the name of Siamese neural networks~\cite{bromley1994signature}. Convolutional counterparts of these networks are also utilized in various image processing tasks~\cite{qi2016sketch}. A more proper name for a Siamese CNN would be a binocular CNN when the input images are stereo images, where each input channel corresponds to an eye. Binocular perspective on CNNs is getting a lot of attention recently~\cite{luo2018object,read2017visual,zeng2019automated}. However, most of these implementations discard the existence of a structure such as the optic chiasma. In other words, a CNN for each eye is usually utilized in parallel in a straightforward way, but no crossing of left and right visual field signals occurs. Hence, this study is novel in that sense as it takes the existence of optic chiasma into account. At this point, the biological functionality of optic chiasma needs to be further investigated as follows.
\begin{figure*}[!h]
	\centering
	\subfloat[CNN]{{\includegraphics[width=5.2cm]{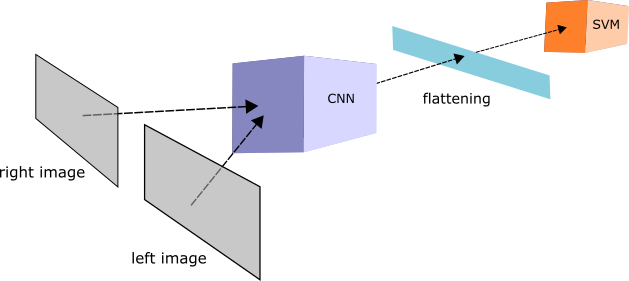} }}%
	\subfloat[BCNN1]{{\includegraphics[width=5.2cm]{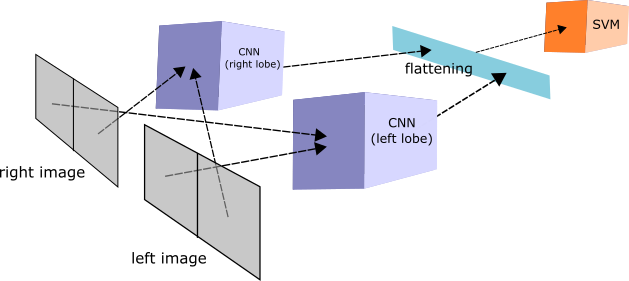} }}%
	\subfloat[BCNN2]{{\includegraphics[width=5.2cm]{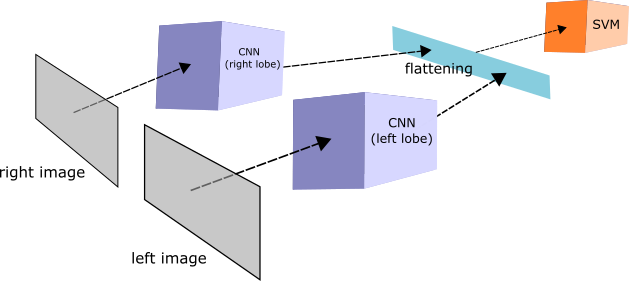} }}%
	\caption{Three architectures considered for stereo datasets. (a) Conventional monocular CNN structure, (b) binocular CNNs structure with regular optic chiasma, (c) binocular CNNs structure without any crossing of signals (the achiasma case).}%
	\label{bcnn_forms}%
\end{figure*}

\subsection{Advantages of Binocular Architecture with Optic Chiasma}

\noindent Most of the attention in this study focuses on the middle structure of Fig.~\ref{fig:example1}, namely the optic chiasma. Optic chiasma allows the signal crossing of the right visual field of right eye and the left visual field of left eye. As a result, right visual fields of both eyes pass through the left brain lobe and left visual fields of both eyes pass through the right brain lobe.

The optic chiasma structure is commonly observed in primates, and mostly in animals with forward facing eyes specifically when fields of view of each eye intersect. Related to this fact, some biological studies indicate the advantage of optic chiasma for eye/hand coordination and depth perception~\cite{larsson2013optic}. Hence, the implementation of such a structure might help robotic studies in which precise manipulation of objects is required in the presence of a stereo camera. Another advantage is that, each eye can individually stimulate both of the brain lobes. Moreover, a certain degree of specialization occurs on left vs. right visual fields since each field is associated with a different brain lobe~\cite{garcea2012right,matthews2015left,nicholls1998contribution}. In a bigger picture, such findings are most probably related to the famous left-right brain paradigm.

Perhaps, more detailed information on the main function of optic chiasma can be gathered by investigating its defects. In albinism, the received visual signals on the retina of each eye are transferred to the opposite brain lobe, namely a full signal crossing occurs instead of a partial one. In achiasma, optic chiasma is totally absent and both left and right visual fields of individual eyes are transferred to the corresponding individual brain lobe without any signal crossing. It is reported that, such abnormalities profoundly affect the organization of visual system but visual perception remains mostly intact~\cite{hoffmann2015congenital}. Although basic visual perception remains intact, various situations requiring different visual and motor abilities must be investigated in order to uncover the true effects of such abnormalities and the advantages of optic chiasma. Nevertheless, a normal brain with a regular optic chiasma has a much high temporal correlation of input signals than the ones have in such abnormalities~\cite{SINHA2012353}.

\subsection{Proposed Binocular Convolutional Neural Networks Architecture}

In this study, advantages of optic chiasma will be investigated through simulations of appropriate CNNs architectures for the classification task. Three different architectures for stereo image datasets have been devised as given in Fig.~\ref{bcnn_forms}. In a conventional monocular CNN architecture, there is no distinction between left and right camera images. For each class, left and right images of the same scene are considered as two distinct training samples (Fig.~\ref{bcnn_forms}(a)). In the first binocular CNNs structure, namely BCNN1 in Fig.~\ref{bcnn_forms}(b), the setup simulates the existence of regular optic chiasma, hence there are two underlying CNNs corresponding to each brain lobe. A CNN is trained for the left visual fields of both left and right images, while considering each visual field as distinct training samples. The other CNN is also trained for the right visual fields of images in the same manner. The second binocular CNNs architecture, namely BCNN2 in Fig.~\ref{bcnn_forms}(c), simulates the achiasma case, in which left and right images are used to train two parallel CNNs without any crossing of signals.

For the above described three architectures, especially involving two different CNNs, the flattened feature outputs of each individual CNN are concatenated in order to simulate a rough functionality of the primary visual cortex. As a final step, these flattened and concatenated features are employed to train and test a support vector machines (SVMs) classifier~\cite{burges1998tutorial} in order to test the classification accuracy.

\subsection{Application to Monocular Image  Datasets}

A crucial point to be pointed out here is that, the proposed binocular perspective naturally suggests that an image dataset is consisting of stereo image pairs. However, an optic chiasma structure can also be implemented for monocular image datasets, while the other way around is impossible for currently available binocular CNN-based approaches in literature. In a straightforward manner, the left and right visual fields of each monocular image could be utilized to train two different CNNs and then simply concatenated with SVMs as in stereo cases. Since this adaptation is out of scope in this study, such extensions of the proposed principle are designated as future work.

\begin{figure*}[!t]
    \centering
    \includegraphics[scale=0.4]{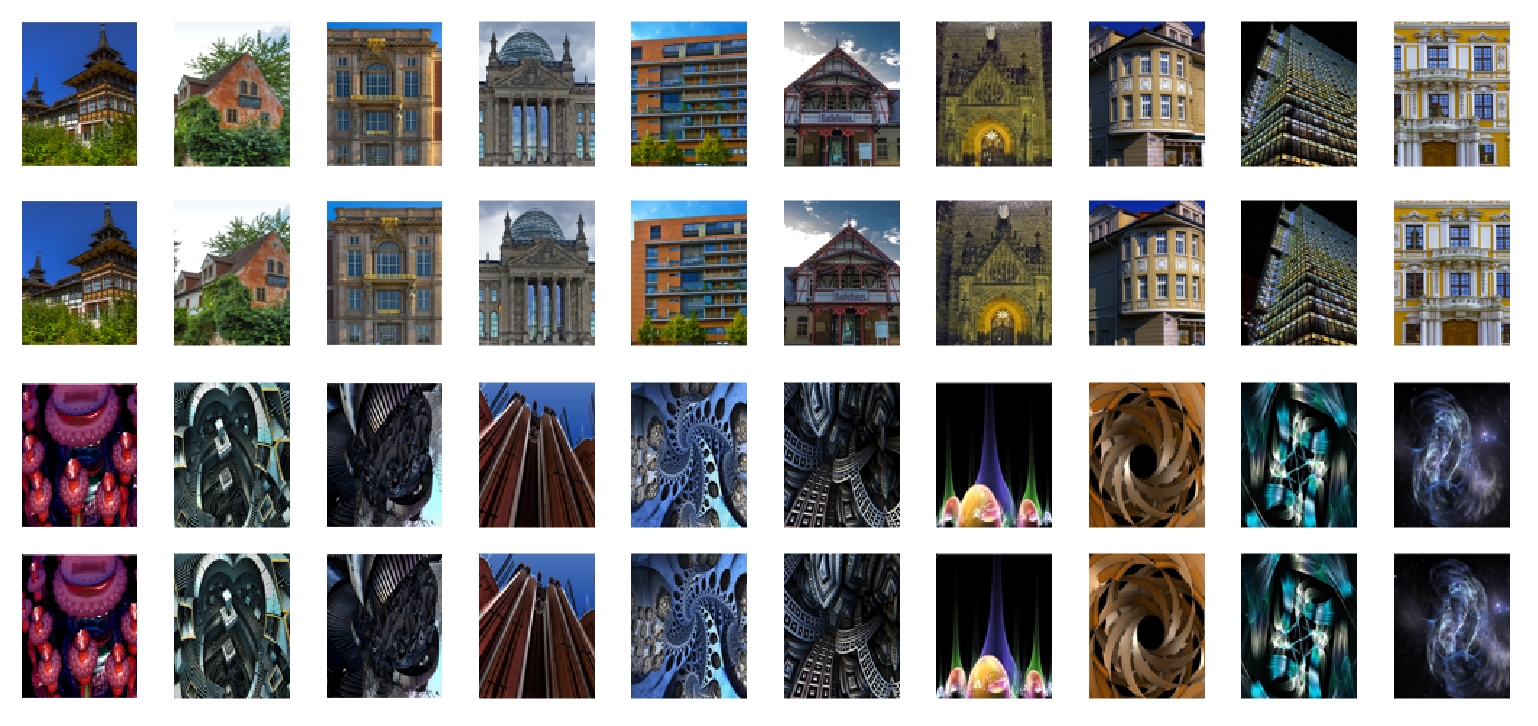}
        \caption{Example stereo image pairs from the dataset. (Top two rows) Class $1$: ``Buildings'' in two views; and (bottom two rows) Class $2$: ``Others'' in two views.}%
    \label{Data-example}%
\end{figure*}

\section{\uppercase{EXPERIMENTAL SETUP and RESULTS}}

Since the starting point of this study depends on the human visual pathway which is binocular, a dataset which consists of stereo images collection is chosen from~\cite{PASSRnet,flickr1024} for the experimental validation. This dataset contains a large-scale of stereo image samples such as animals, humans, buildings, synthetics, plants and sculptures. To perform experiments and compare the classification results, $100$ images for Class $1$: ``Buildings'' and  $100$ images for Class $2$: ``Others'' are selected. Note that the selected images from Class $2$ are very challenging in which some of them are very similar images of buildings. Some examples of image pairs are illustrated in Fig~\ref{Data-example}. To avoid high computational cost and heterogeneous spatial resolution while the experimentation process, the dataset images are resized to $400\times 500\times 3$ pixels of uniform resolution. In order to increase the number of images then, data augmentation is applied by reflection (both horizontally and vertically), rescaling (upsampling/downsampling and downsampling/upsampling), random noise distortion with small variances, rotation in very small angles and translation in very small ranges (both horizontally and vertically).
\begin{figure*}[!t]
	\centering
	\includegraphics[width=14cm]{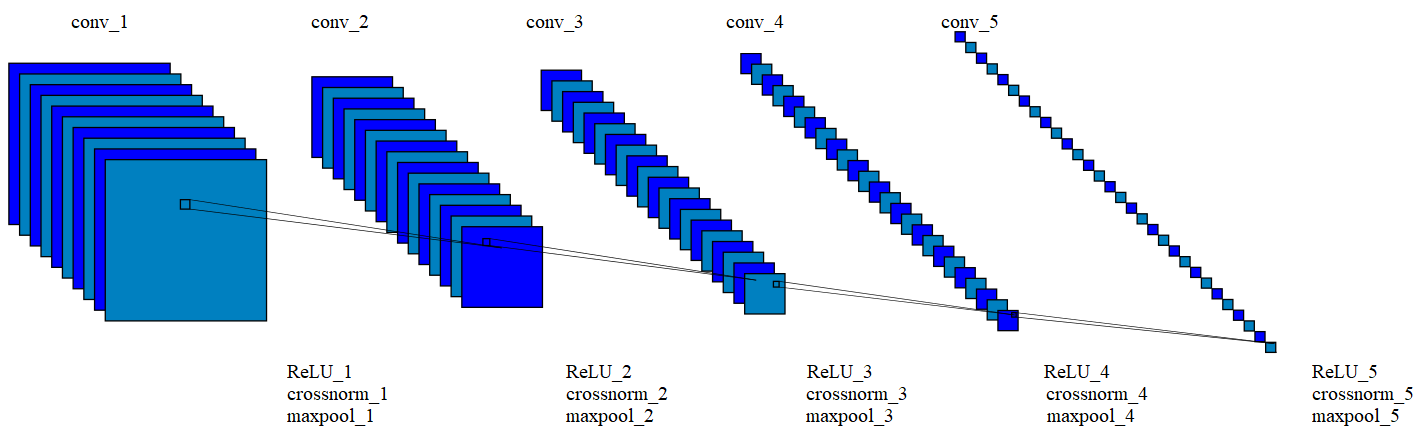}
	\caption{The CNNs architecture utilized in the experiments. Feature outputs in $\texttt{maxpool\_5}$ are used in SVMs classifier.}%
	\label{CNN-arch}%
\end{figure*}

In the training phase, a successful $24$-layer CNN architecture is adapted from~\cite{turkanEBBT,turkanASYU}. This structure consists in an input layer followed by; a convolution layer, a ReLU layer, a cross-normalization layer and a max-pooling layer repeated in $5$ distinct blocks in the given order (Fig.~\ref{CNN-arch}); a fully-connected layer (binary output), a softmax layer and a final classification layer. The input layers are of size $400\times 500\times 3$ for CNN and BCCN2. On the other side, the simulated left and right visual field inputs for BCNN1 are of size $200\times 500\times 3$, in other words images are halved and their left and right parts are obtained for left and right visual fields respectively. The convolutional filter parameters are $15\times 15\times 10$ for $\texttt{conv\_1}$, $11\times 11\times 15$ for $\texttt{conv\_2}$, $9\times 9\times 20$ for $\texttt{conv\_3}$, $7\times 7\times 25$ for $\texttt{conv\_4}$, $5\times 5\times 30$ for $\texttt{conv\_5}$. All ReLU layers are followed by cross-normalization layers to express more importance to large activations in certain neighborhoods. Finally, max-pooling layers operate on $3\times 3$ regions with stride $2$.

The augmented stereo dataset is randomly divided into a train set ($60\%$ of the dataset) and a test set ($40\%$ of the dataset). Three CNNs architectures in Fig.~\ref{bcnn_forms} are learned with respect to their (binocular) structures using the same training set of images. After the training process of all CNNs architectures, the flattened feature outputs in $\texttt{maxpool\_5}$ (concatenated in BCNN1 and BCNN2) are fed to a binary SVMs learner (a one-versus-one coding design) per CNNs architecture. In the testing phase then, the test set images (which are not utilized during training) are used to produce the flattened feature outputs in $\texttt{maxpool\_5}$ (concatenated in BCNN1 and BCNN2) and classified with the respective SVMs classifier. 

The aforementioned procedure is repeated for $25$ independent simulations with random initializations and randomly constructed training and test sets to measure the consistency of three CNNs systems. The obtained minimum, maximum and average classification results are reported in Table~\ref{tab:table-rates}. The achiasmatic structure, namely BCNN2, performs better than the monocular CNN architecture at both maximum and minimum. However, it has the lowest performance on average, thus it is not robust in case of random initializations. It is apparent that BCNN1 simulating optic chiasma provides the best performance in this classification task. Not only its classification accuracy rates are highest in all cases, but also the standard deviation of these results is the lowest. One can conclude that BCNN1 is also the most robust structure in case of random initializations. The optic chiasma structure in BCNN1 increases the classification accuracy approximately $2\%$ on average.
\begin{table}[!t]
	\centering
	\small
	\caption{Experimental results obtained from $25$ independent simulations with random initializations.}
	{%
		\begin{tabular}{|c|c|c|c||c|}
			\hline
			\multicolumn{1}{|l|}{} &      min (\%)     &       max (\%) &   mean (\%)    &  stdev \\ \hline
			CNN                    &       $84.38$      &    $97.50$    &   $94.43$      &  $0.0248$    \\ \hline
			BCNN1                  &       $92.50$      &    $99.38$    &   $96.28$      &  $0.0174$    \\ \hline
			BCNN2                  &       $90.63$      &    $98.12$    &   $93.78$      &  $0.0221$    \\ \hline
	\end{tabular}}
	\label{tab:table-rates}
\end{table}

\section{\uppercase{Conclusion}}

In summary, forward facing eyes, in other words having intersecting fields of view, evolutionarily lead to the development of a structure called optic chiasma. Such structure allows right visual fields of both eyes to pass through the left brain lobe and left visual fields of both eyes to pass through the right brain lobe, later to be processed in the primary visual cortex. Naturally, this biological brain architecture has a much high temporal correlation of input signals than the ones lacking optic chiasma. Artificial simulations on a challenging image dataset prove its superiority over monocular or achiasmatic alternatives. Such superior results should not be surprising in a corresponding setting where fields of views of images intersect by replicating optic chiasma.

As an extension of this study, optic chiasma principle may be applied on monocular image datasets as mentioned earlier. Another extension can be to investigate the nature of convolutional filters appearing in left vs. right CNNs within the BCNN1 structure on different visual tasks to explain whether respective visual fields of each eye and the associated brain lobe are responsible for different learning abilities of the same scene. Further studies on this issue will pave way to better understanding of visual processing systems in general.

\bibliographystyle{apalike}
{\small

\bibliography{ref}}

\vfill
\end{document}